\documentclass[10pt,twocolumn,letterpaper,table]{article}

\usepackage[pagenumbers]{cvpr} 

\usepackage{xcolor}
\definecolor{cvprblue}{rgb}{0.21,0.49,0.74}
\definecolor{lightgray}{rgb}{0.83, 0.83, 0.83}
\definecolor{greensup}{rgb}{0.25, 1, 0.4}
\definecolor{redsup}{rgb}{0, 1, 0.6}
\definecolor{lightblue}{rgb}{0.85, 0.95, 1}
\definecolor{lightorange}{rgb}{1, 0.95, 0.85}
\definecolor{lightpink}{rgb}{1, 0.9, 0.95}
\definecolor{indiagreen}{rgb}{0.07, 0.53, 0.03}
\definecolor{lightcream}{rgb}{0.98, 0.96, 0.90}
\definecolor{beige}{rgb}{0.96, 0.96, 0.86} 
\definecolor{emerald}{rgb}{0.31, 0.78, 0.47}
\definecolor{lightgoldenrodyellow}{rgb}{0.98, 0.98, 0.82}
\definecolor{blue}{rgb}{0.0, 0.0, 1.0}
\definecolor{pastelyellow}{rgb}{0.99, 0.99, 0.59}
\definecolor{mygreen}{RGB}{0,128,0}
\definecolor{myred}{rgb}{1.0, 0.03, 0.0}
\newcommand{\hlc}[2][yellow]{{%
    \colorlet{foo}{#1}%
    \sethlcolor{foo}\hl{#2}}%
}
\usepackage{arydshln}
\usepackage{amsmath}
\usepackage{amssymb}
\usepackage{pifont}
\usepackage{mathtools}
\usepackage{booktabs}
\usepackage{float}
\usepackage{multirow}
\usepackage{algorithm}
\usepackage{algpseudocode}
\usepackage{enumitem}%
\usepackage{multibib}
\usepackage{soul}
\usepackage{fontawesome}
\usepackage{tcolorbox}
\usepackage{adjustbox}
\newcommand{\cmark}{\ding{51}}%
\newcommand{\xmark}{\ding{55}}%
\newcommand{\changeurlcolor}[1]{\hypersetup{urlcolor=#1}}    

\algrenewcommand\algorithmicrequire{\textbf{Input:}}
\algrenewcommand\algorithmicensure{\textbf{Output:}}

\usepackage{listings}

\usepackage[pagebackref,breaklinks,colorlinks]{hyperref}

\def\paperID{0003} 
\def\confName{CVPRW}
\def\confYear{2025}

\title{NeIn: Telling What You Don't Want}
\author{\fontsize{11.5}{13}\selectfont Nhat-Tan Bui$^{1}$, Dinh-Hieu Hoang$^{2}$, Quoc-Huy Trinh$^{3, 4}$, 
Minh-Triet Tran$^{2}$, Truong Nguyen$^{5}$, Susan Gauch$^{1}$ \\
\normalsize
$^1$University of Arkansas, USA \enspace
$^2$University of Science, VNU-HCM, Vietnam \\
\normalsize
$^3$Aalto University, Finland \enspace
$^4$SpexAI GmbH, Germany \enspace
$^5$University of California, San Diego, USA \\
\normalsize \url{https://tanbuinhat.github.io/NeIn/}
}

\begin{document}
\maketitle
\begin{abstract}

Negation is a fundamental linguistic concept used by humans to convey information that they do not desire. Despite this, minimal research has focused on negation within text-guided image editing. This lack of research means that vision-language models (VLMs) for image editing may struggle to understand negation, implying that they struggle to provide accurate results. One barrier to achieving human-level intelligence is the lack of a standard collection by which research into negation can be evaluated.
This paper presents the first large-scale dataset, \textbf{Ne}gative \textbf{In}struction (NeIn), for studying negation within instruction-based image editing. Our dataset comprises \textbf{366,957 quintuplets}, i.e., source image, original caption, selected object, negative sentence, and target image in total, including \textbf{342,775 queries for training} and \textbf{24,182 queries for benchmarking} image editing methods.
Specifically, we automatically generate NeIn based on a large, existing vision-language dataset, MS-COCO, via two steps: generation and filtering. 
During the generation phase, we leverage two VLMs, BLIP and InstructPix2Pix (fine-tuned on MagicBrush dataset), to generate NeIn's samples and the negative clauses that expresses the content of the source image. In the subsequent filtering phase, we apply BLIP and LLaVA-NeXT to remove erroneous samples.
Additionally, we introduce an evaluation protocol to assess the negation understanding for image editing models.
Extensive experiments using our dataset across multiple VLMs for text-guided image editing demonstrate that even recent state-of-the-art VLMs struggle to understand negative queries.

\end{abstract}    
\section{Introduction}
\label{sec:intro}

When it comes to training vision-language models (VLMs), we have to consider a wide range of human information needs, requiring systems to handle a wide range of user-generated queries. They range from simple and straightforward ones like “describe the image" to complex prompts involving rich contextual detail and creative reasoning.

Inspired by Laurence R. Horn \cite{negation}, \textit{“negation is a sine qua non of every human language but is absent from otherwise complex systems of animal communication.”}  In this work, we address the problem of negative queries that specify information that should be excluded, a ubiquitous feature in human language. Examples of negative queries in image editing tasks include “The bathroom area \textit{without} a curtain" or “The street with a person, but \textit{not} with a car."

\begin{figure}[t]
\begin{center}
\includegraphics[width=\linewidth]{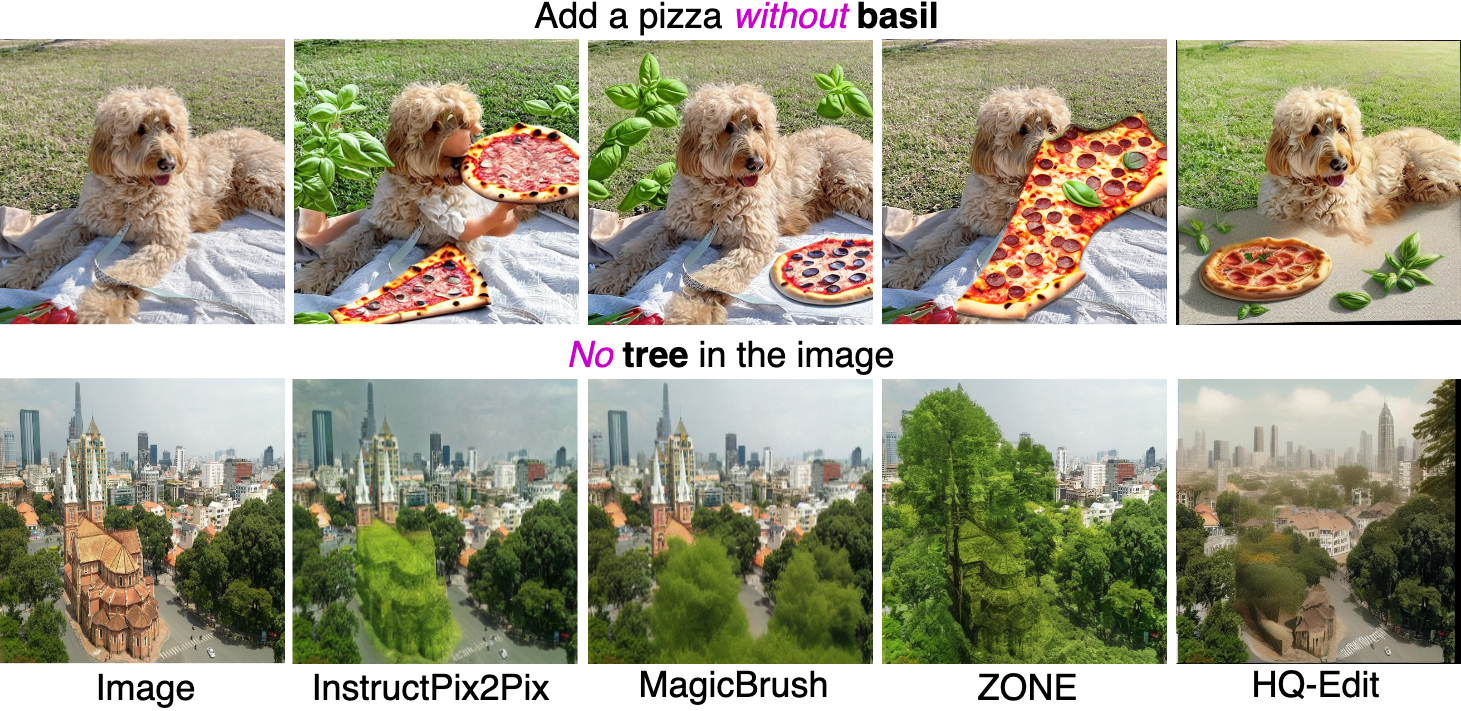}
\vspace{-6mm}
\caption{The failures of recent text-guided image editing methods in understanding the negative queries.}
\label{fig:motivation}
\end{center}
\vspace{-8mm}
\end{figure}

\begin{table*}[t]
    \centering
    \resizebox{\linewidth}{!}{
    \begin{tabular}{l|c|c|c|c|c|c|c}
    \toprule
    \multirow{2}{*}{Datasets} & \multirow{2}{*}{Tasks} & \multicolumn{2}{c|} {Train} & \multicolumn{2}{c|} {Validation} & \multicolumn{2}{c} {Total} \\
    \cline{3-8}
    & & \#Negative & \#All & \#Negative & \#All & \#Negative & \#All \\
    \hline
    CC12M \cite{cc12m} & Pre-training  & -- & -- & -- & -- & 314,181 (\textcolor{red}{2.53\%}) & 12,423,374\\
    LAION-400M \cite{laion} & Pre-training  & -- & -- & -- & -- & 2,404,784 (\textcolor{red}{0.58\%}) & 413,862,224\\
    MS-COCO'14 \cite{coco} & Image Captioning & 1,761 (\textcolor{red}{0.43\%})   & 414,113 & 886 (\textcolor{red}{0.44\%}) & 202,654 & -- & 616,767\\
    SBU Captions \cite{sbu} & Image Captioning & -- & -- & -- & -- & 26,222 (\textcolor{red}{2.62\%}) & 1,000,000 \\
    CC3M \cite{cc3m} & Image Captioning & 54,219 (\textcolor{red}{1.63\%}) & 3,318,333 & -- & -- & -- & 3,369,218 \\
    CIRR \cite{cirr} & Composed Image Retrieval & 868 (\textcolor{red}{3.08\%}) & 28,225 & 130 (\textcolor{red}{3.11\%}) & 4,181 & -- & 36,554  \\
    InstructPix2Pix\cite{instructpix2pix} & Image Editing & 77 (\textcolor{red}{0.02\%}) & 313,010 & -- & -- & -- & 313,010 \\
    MagicBrush \cite{magicbrush} & Image Editing & 54 (\textcolor{red}{0.61\%}) & 8,807 & 6 (\textcolor{red}{1.17\%}) & 528 & -- & 10,388 \\
    \bottomrule
    \end{tabular}}
    \caption{Statistic of captions in current image-caption pair datasets. We identify negative captions by 32 negative words: “no", “not", “without", “don't", “doesn't", “never", “none", “neither", “nothing", “can't", “isn't", “aren't", “didn't", “did not", “isn't", “is not", “aren't", “are not", “wasn't", “was not", “weren't", “were not", “won't", “will not", “hasn't", “has not", “haven't", “have not", “can't", “can not", “couldn't", and “could not".}
    \label{tab:statistic}
\vspace{-4mm}
\end{table*}

Research that explicitly tackles the negation problem for neural networks has mainly focused on natural language understanding \cite{ravichander2022condaqa, truong2023language, weller2023nevir, zhang2024excluir, mai2024bert}, and a few vision-language tasks including video retrieval \cite{wang2022learn}. However, in image editing,  many recent SOTAs such as InstructPix2Pix \cite{instructpix2pix}, MagicBrush \cite{magicbrush}, ZONE \cite{zone}, and HQ-Edit \cite{hqedit} fail to understanding negative queries, as Figure \ref{fig:motivation} illustrates. In contrast, the models seem to focus on adding objects that need to be excluded from the input images (for the query \textit{no tree in the image}, more trees are added, occluding the cathedral).

Although image editing VLMs can be prompted with instructions such as “remove," understanding negation is crucial for these models to achieve human-level intelligence. This is because humans frequently use negation in various ways, and not every negative cue can simply be replaced by using “remove."

One possible reason why VLMs fail to understand negation is the lack of negative descriptions in current image-caption pair datasets, e.g., MS-COCO \cite{coco}, SBU Captions \cite{sbu}, CC12M \cite{cc12m}, LAION-400M \cite{laion}, etc. Since the nature of captions in these datasets is to describe objects and visual concepts in the image as well as the stories related to them, they lack negative clauses.
As illustrated in Table \ref{tab:statistic}, the number of negative sentences in image-caption pair datasets is very small. Furthermore, some captions contain negative words but they actually do not describe what is not present in the image, e.g., “do not spend money in stores with this sign," “things you cannot miss."
This leads to two consequences: (1) VLMs do not understand the meaning of negative words because of the heavily-biased dataset during training, and (2) there is no evaluation data to assess the capability of VLMs in understanding negation. 

To tackle this issue, we present a pipeline for constructing a new dataset, \textbf{Ne}gative \textbf{In}struction (NeIn\footnote{Nein means “no" in German.}), designed for training and evaluating VLMs on negation understanding, specifically within the context of image editing. Particularly, we use BLIP \cite{blip} to identify objects that are not present in the source image. We expand the captions that represent the content of source image by incorporating negative words for objects that are not present. One example of a negative sentence would be “The image \textit{doesn't} have an \textit{apple}". We then generate counter-example target images that contain those absent objects using InstructPix2Pix \cite{instructpix2pix} fine-tuned on MagicBrush \cite{magicbrush}. We retain only acceptable-quality images by filtering target samples using BLIP and LLaVA-NeXT \cite{llavanext} to remove images that have been excessively altered from the original content of the source image or that make the excluded object unrecognizable. Thus, by combining the target NeIn sample with the corresponding negative query, we can obtain the source image.


In order to evaluate the performance of VLMs on our dataset, we propose an evaluation method for image editing models to assess both the ability to remove objects from images in response to negative clauses and the ability to retain the original objects in the image after modification by the queries. To summarize, our main contributions are:
\begin{itemize}[noitemsep,topsep=0pt]
\itemsep0em 
    \item We investigate the ability of VLMs to interpret negation cues in text-guided image editing, leading to the creation of the first large-scale vision-language negation dataset for this task, termed NeIn.
    \item We introduce a pipeline to generate NeIn, an extensive dataset comprising 366,957 quintuplets. This dataset focuses on the understanding of negation, a fundamental linguistic concept, for image editing VLMs.
    \item We propose an evaluation method for negation understanding that can be used by future researchers. Using our evaluation method, we observe that VLMs in image editing task have difficulty  comprehending negative instructions. This insight opens a new research direction for improving negation understanding for VLMs.
    
\end{itemize}
\section{Related Work}
\label{sec:related}


\begin{figure*}
\begin{center}
\includegraphics[width=\linewidth]{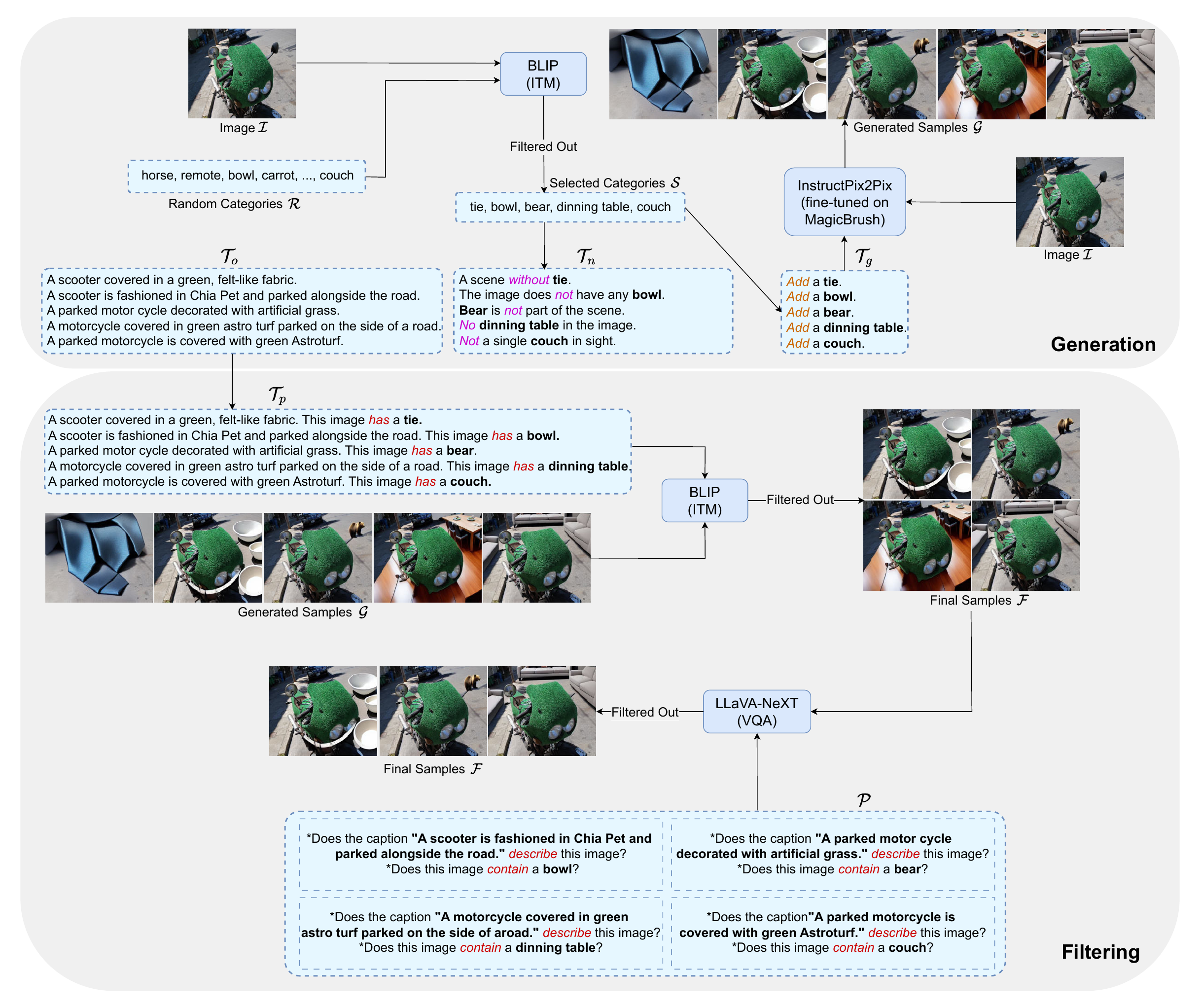}
\caption{The process to create our dataset. It consists of two main steps: generation and filtering. ITM and VQA are image-text matching and visual question answering, respectively.}
\label{fig:process}
\end{center}
\vspace{-7mm}
\end{figure*}



\textbf{Negation Understanding}, a major linguistic topic, is becoming prominent in research. In natural language understanding, Ravichander et al. proposed CondaQA \cite{ravichander2022condaqa} which is a question answering dataset specifically designed for negation understanding. Experiments conducted on CondaQA revealed that deep learning models may have a simple trick that they reverse the rank list when they see the negation cues, leading to acceptable results when encountering fully negated queries. Nevertheless, their performance still suffer a severe degradation on composed queries. Truong et al. \cite{truong2023language} investigated the LLMs' ability to understand negation under various settings. NevIR \cite{weller2023nevir} benchmarked models against a simple yet brilliant task, ranking two paragraphs regarding a question that is relevant to only one of them. Measured in pair-wise accuracy, volunteers easily achieved a perfect score of 100\%, far superior to most of the models whose performances are below 25\%. The best ones, using cross encoders, scored below 50\%. ExcluIR \cite{zhang2024excluir} included an evaluation benchmark comprising 3,452 manually curated queries, along with a training set of 70,293 queries with a positive document and a negative document. The authors have concluded that even after fine-tuned on negated dataset, all models still lag behind human performance a great deal. SetBERT \cite{mai2024bert} recently proposed fine-tuning on a synthesized dataset with a focus on predicting negative samples using inverse-contrastive learning, while BoolAttn \cite{mai2025boolean} took a different approach by directly adjusting attention scores to downweight tokens affected by negation.
In conclusion, all the results unanimously show that even SOTAs struggle to comprehend negation cues.

For visual data retrieval, Wang et al. \cite{wang2022learn} measured models on original dataset MSR-VTT3k, its negated version, and composed version. However, models can cheat on the negated version by inverting the result on the original dataset ($\Delta$Recall and $\Delta$MIR are nearly zero) but on composed queries, all the models perform very poorly (Recalls at rank N=1, 10 are respectively less than 12\% and 46\%, and Mean Inverted Rank is less than 0.23), lagging behind their own performance on the original dataset by a large margin.

Similar to our work, Singh et al. \cite{singh2024learn} recently introduced the CC-Neg dataset tailored for negation problem, contains 228,246 images, true captions and negated captions. However, the negated captions of CC-Neg falsely describe their corresponding images.
More precisely, the authors generated the negated captions so that they and their images can be used as negative pairs in the contrastive loss to finetune the vision-language pre-trained models, whereas in our dataset NeIn, the negated captions still describe exactly their original images. 
Other differences will be clarified in the Sections \ref{sec:method} and \ref{sec:exp}.


\begin{table*}[t]
    \centering
    \renewcommand{\arraystretch}{1.2}
    \setlength{\tabcolsep}{2pt}
    \resizebox{\linewidth}{!}{
    \begin{tabular}{l|l|l|l}
    \toprule
    1) The image \textit{doesn't} have any \{$\mathcal{S}$\}.
 & 2) \{$\mathcal{S}$\} is \textit{not} part of the scene.
 & 3) \textit{No} \{$\mathcal{S}$\} present in the image. & 4) The image is \textit{without} \{$\mathcal{S}$\}.  \\
 \hline
    5) The image does \textit{not} have any \{$\mathcal{S}$\}. & 6) The image \textit{lacks} \{$\mathcal{S}$\}. & 7) \textit{No} \{$\mathcal{S}$\} in the image. & 8) A scene \textit{without} \{$\mathcal{S}$\}. \\
     \hline
    9) The image \textit{cannot} have any \{$\mathcal{S}$\}. & 10) \textit{Not} a single \{$\mathcal{S}$\} in sight. & 11) \{$\mathcal{S}$\} is \textit{missing} from the image. & 12) The image \textit{lacks} the presence of \{$\mathcal{S}$\}.\\
     \hline
    \multicolumn{4}{c} {
    13) \{$\mathcal{S}$\} is \textit{nowhere} to be seen in the image.}\\
    \bottomrule
    \end{tabular}}
    \caption{The pre-defined formats for $\mathcal{T}_{n}$. These formats contain the negative words: “no", “not", “without", “doesn't", “does not", “cannot", “lacks", “missing", and “nowhere".}
    \vspace{-4mm}
    \label{tab:t_negative}
\end{table*}

\section{NeIn Dataset}
\label{sec:method}

NeIn is designed specifically to address the challenge of negation understanding in VLMs for image editing. Each tuple in NeIn consists of a natural image $\mathcal{I}$ and its original caption $\mathcal{T}_o$, selected object $\mathcal{S}$,  sentence containing negative clause $\mathcal{T}_n$ and a synthetic image $\mathcal{G}$ that satisfies the original caption but not its corresponding negative sentence.

The creation of NeIn involves two primary stages: the first stage is \textit{generation}, which employs BLIP \cite{blip} and InstructPix2Pix \cite{instructpix2pix}, fine-tuned on MagicBrush\cite{magicbrush}, to generate target samples; the second stage is \textit{filtering}, where BLIP and LLaVA-NeXT \cite{llavanext} are utilized to remove erroneous samples.
\setlength{\textfloatsep}{2pt}
\subsection{Generation Pipeline}
\label{sec:generation}
The data generation process of NeIn is shown in upper part of Figure \ref{fig:process}. The main idea is that given image $\mathcal{I}$ and a corresponding caption $\mathcal{T}_{o}$ describing what objects are present in $\mathcal{I}$, we will find a negative clause, termed $\mathcal{T}_{n}$, such that it satisfies the content of source image $\mathcal{I}$. In our case, $\mathcal{I}$ and $\mathcal{T}_{o}$ are the image and its corresponding captions from MS-COCO \cite{coco}.
More precisely, for each tuple, we arbitrarily select a collection of 15 object categories $\mathcal{R}$ from MS-COCO. This means the categories vary for each tuple. Then, we find out which objects in $\mathcal{R}$ do not appear in $I$ using the image-text matching (ITM) from BLIP \cite{blip}. Note that the original caption $\mathcal{T}_{o}$ only describes the main objects and their relationships within the source image $\mathcal{I}$; it does not account for all the categories present in $\mathcal{I}$. For instance, $\mathcal{T}_{o}$ is “bedroom scene with a bookcase, blue comforter, and window," but $\mathcal{I}$ also includes a mirror, trees, a lamp, and a cabinet. Therefore, using BLIP instead of identifying objects in $\mathcal{T}_{o}$ to evaluate $\mathcal{R}$ will reduce errors for NeIn's samples. We choose BLIP over other VLMs because it is sufficiently fast to process large quantities of images. We accept false positive results (when an object is present in the image but BLIP fails to detect it) during the generation step. If an object is too small or distorted to the point where BLIP cannot recognize it, we consider that object is not present in the image.

\begin{algorithm}[ht]
\textbf{Input:} \\
$\mathcal{I}$: source image \\
$\mathcal{R}$: list of 15 random object categories \\
$\mathcal{P}$: list of pre-defined formats for negative clauses

\textbf{Output:}\\
$\mathcal{S}$, $\mathcal{T}_{n}$, $\mathcal{T}_{g}$, $\mathcal{G}$: lists of selected object categories, negative sentences, prompts, and their corresponding synthetic images, respectively

\caption{Generation}
\label{alg:generation}
\begin{algorithmic}[1]
\State $\mathcal{S} \coloneqq$ [], $\mathcal{T}_{n} \coloneqq$ [], $\mathcal{T}_{g} \coloneqq$ [], $\mathcal{G} \coloneqq$ []  

\For{each object $\mathcal{R}^{(i)}$ in $\mathcal{R}$}
\Comment{Categories not in $I$}
    \State $c^{(i)} \gets$ ITM($\mathcal{I}$, $\mathcal{R}^{(i)}$) 
    \Comment{Cosine similarity}
    \Statex \hspace{0.45cm} \textcolor{teal}{\# threshold $\alpha = 0.4$}

    \If{$c^{(i)} < \alpha$}  \Comment{Add categories}
        \State append ($\mathcal{R}^{(i)}$, $c^{(i)}$) to $\mathcal{S}$
    \EndIf
\EndFor

\State $\mathcal{S} \gets$ top$(\mathcal{S}, 5)$  
\Comment{Select 5 lowest-cosine categories}
\For{each object $\mathcal{S}^{(i)}$ in $\mathcal{S}$} \Comment{Form sentences}
    \State $p \gets$ random$(\mathcal{P})$ \Comment{Choose a random format}
    
    \State $\mathcal{T}_n^{(i)} \gets$ gen$(p, \mathcal{S}^{(i)})$  
    \Comment{Negative sentence}
    \State append $\mathcal{T}_n^{(i)}$ to $\mathcal{T}_{n}$
    \State $\mathcal{T}_g^{(i)} \gets$ “Add a/an \{$\mathcal{S}^{(i)}$\}"
    \Comment{Instruction sentence}
    \State append $\mathcal{T}_g^{(i)}$ to $\mathcal{T}_{g}$
\EndFor

\For{each prompt $\mathcal{T}_g^{(i)}$ in $\mathcal{T}_{g}$} \Comment{NeIn samples gen}

    \State $\mathcal{G}^{(i)} \gets$ generator$(\mathcal{I}, \mathcal{T}_g^{(i)})$  
    \Comment{Generated sample}
    \State append $\mathcal{G}^{(i)}$ to $\mathcal{G}$
\EndFor
\State \Return $\mathcal{S}, \mathcal{T}_{n}, \mathcal{T}_{g}, \mathcal{G}$
\end{algorithmic}
\end{algorithm}

We select 5 objects with lowest scores and generate their corresponding negative sentences based on some pre-defined formats, illustrated in Table \ref{tab:t_negative}. The negative sentences include negative words such as “doesn't," “not," “without" and $\mathcal{S}$ represents the selected objects, indicating that these objects are not present in the image $\mathcal{I}$. Concatenating $\mathcal{T}_{n}$ with the original caption $\mathcal{T}_{o}$, we attain a new caption that still matches the original image $\mathcal{I}$.

Next, our goal is to create an image $\mathcal{G}$ that $\mathcal{T}_{o}$ matches it but not $\mathcal{T}_{n}$, which means the object specified in $\mathcal{T}_{n}$ is present in $\mathcal{G}$.
Theoretically, we can create $\mathcal{G}$ by any image editing method adding those objects to the source image. In the case of NeIn, we choose a version of InstructPix2Pix \cite{instructpix2pix}, a diffusion-based deep neural network, fine-tuned on MagicBrush \cite{magicbrush} due to the high quality of its outputs. Thus, in the context of image editing, given image $\mathcal{G}$, $\mathcal{T}_{n}$ will be a query for removing some object in $\mathcal{G}$, taking $\mathcal{I}$ as one of the best results. For instance, if $\mathcal{I}$ is a picture of a dog in a garden, $\mathcal{T}_{n}$ could be “the photo does not have any laptop," and $\mathcal{G}$ would be a picture of a dog and a laptop in the garden. To summarize, Algorithm \ref{alg:generation} illustrates the steps to accomplish the generation phase. Note that each image in MS-COCO has 5 captions, so we can generate 5 tuples for each image.

\subsection{Filtering Pipeline}
\label{sec:filtering}
The data filtering process of NeIn is shown in lower part of Figure \ref{fig:process}. The main purpose of this stage is to eliminate images that significantly alter the content of query image $\mathcal{I}$ or difficultly identify object categories $\mathcal{S}$. The specific examples are the $1^{\text{st}}$ and $4^{\text{th}}$ images in $\mathcal{G}$ (Figure \ref{fig:process}), where the image is completely transformed and no longer retains the original objects like the motorcycle and the green fabric. If the original content is no longer preserved or the object $S$ is hard to distinguish, then, when combined with $\mathcal{T}_{n}$, the model can no longer output $\mathcal{I}$.

We employ a two-stage filtering strategy: the first stage leverages the image-text matching (ITM) function from BLIP \cite{blip}, while the second utilizes visual question answering (VQA) from LLaVA-NeXT \cite{llavanext}. We select these models because we need to verify two aspects: original captions and selected object categories. Using ITM (represented by BLIP) and VQA (represented by LLaVA-NeXT) enable us to design prompts for checking these two aspects.

In order to perform the first filtering stage, we generate $\mathcal{T}_{p}$ by combining $\mathcal{T}_{o}$ with a prompt “This image \textit{has} a/an \{$\mathcal{S}$\}", where $\mathcal{S}$ represents the selected object category. We utilize BLIP to calculate the matching score of $\mathcal{T}_{p}$ and the generated samples $\mathcal{G}$ to eliminate the erroneous samples from $\mathcal{F}$.

Results from the first stage are further filtered in the second stage to ensure the quality of NeIn's samples. For this second stage, we design two prompts $\mathcal{P}$: (1) Does the caption $\mathcal{T}_{o}$ \textit{describe} this image? and (2) Does this image \textit{contain} a/an \{$\mathcal{S}$\}? The first prompt ensures that the content of the samples $\mathcal{F}$ does not deviate significantly from $\mathcal{I}$, while the second prompt checks whether the selected objects appear in $\mathcal{F}$. We require LLaVA-NeXT to output “Yes" or “No" for both prompts, and if either prompt returns “No," we remove that sample from $\mathcal{F}$. Algorithm \ref{alg:filtering} shows the pseudo code for our filtering phase. The examples of NeIn's sample after filtering step and the data statistics are shown in the supplementary material.

\begin{algorithm}[h]
\textbf{Input:} \\
$\mathcal{G}$, $\mathcal{T}_o$: generated images and its original caption \\
$\mathcal{S}$: list of selected object categories\\
\textbf{Output:} \\
$\mathcal{F}$: final samples of NeIn
\caption{Filtering}
\label{alg:filtering}
\begin{algorithmic}[1]
\State $\mathcal{F} \coloneqq []$
\Statex  \textcolor{teal}{\# first filtering stage by ITM}
\For{each tuple $(\mathcal{S}^{(i)}, \mathcal{T}_o^{(i)}, \mathcal{G}^{(i)})$ in $(\mathcal{S}, \mathcal{T}_o, \mathcal{G})$}
    \State $\mathcal{T}_p^{(i)} \gets \mathcal{T}_o^{(i)}$ + “This image has a/an \{$\mathcal{S}^{(i)}$\}"
    \State $c^{(i)} \gets$ ITM($\mathcal{G}^{(i)}$, $\mathcal{T}_p^{(i)}$) \Comment{Cosine similarity}
    \Statex \hspace{0.45cm} \textcolor{teal}{\# threshold $\alpha = 0.4$}
    \If{$c^{(i)} > \alpha$}  \Comment{Add samples}
        \State append $\mathcal{G}^{(i)}$ to $\mathcal{F}$
    \EndIf
\EndFor

\Statex  \textcolor{teal}{\# second filtering stage by VQA}

\For{each tuple $(\mathcal{S}^{(i)}, \mathcal{T}_o^{(i)}, \mathcal{F}^{(i)})$ in $(\mathcal{S}, \mathcal{T}_o, \mathcal{F})$}
\Statex \hspace{0.45cm} \textcolor{teal}{\# two pre-defined prompts}
    \State $\mathcal{P}_1^{(i)} \gets$ “Does the caption \{$\mathcal{T}_o^{(i)}$\} describe this image?"
    \State $\mathcal{P}_2^{(i)} \gets$ “Does this image contain a/an \{$\mathcal{S}^{(i)}$\}?"  
    \State $b_1^{(i)} \gets$ VQA($\mathcal{F}^{(i)}$, $\mathcal{P}_1^{(i)}$) \Comment{Boolean result}
    \State $b_2^{(i)} \gets$ VQA($\mathcal{F}^{(i)}$, $\mathcal{P}_2^{(i)}$) \Comment{Boolean result}

    \If{$b_1^{(i)} = \text{“No"}$ \textbf{or} $b_2^{(i)} = \text{“No"}$}
    \Comment{Filter out}
        \State remove $\mathcal{F}^{(i)}$ from $\mathcal{F}$
    \EndIf
\EndFor
\State \Return $\mathcal{F}$
\end{algorithmic}
\end{algorithm}

\subsection{Discussion} 
First, in contrast to CC-Neg \cite{singh2024learn} whose tuples  contain only a natural image, its corresponding caption, and a negated caption, we design NeIn so that for every caption, there is at least one target image associated with it. 
As a consequence, NeIn, with \textit{a source image, corresponding captions, selected objects, negative sentences, and target images}, may be suitable for tasks that require target images (e.g., image editing, composed image retrieval, visual grounding), thus potentially having more applications than CC-Neg. Besides, CC-Neg only consists of basic negative words like “no", “not", “without"; while NeIn is more diverse.

Second, this dataset is intended to support research on negation understanding, as a purely mathematical logic,  rather than generating realistic images, which is related to naturalness. That means, “a carrot within a traffic scene" may seem absurd but humans can reliably and effortlessly determine whether there is a carrot in the photo regardless of other aspects of the photo. Therefore, the fact that NeIn's samples are synthetic does not impact the overall quality of the dataset. In other words, we want to assess the models' ability to answer logical questions, unaffected by irrelevant factors such as naturalness, context, artistic style.

Third, one concern is raised by how we generate images corresponding to $\mathcal{G}$. Generated by the image editing diffusion model InstructPix2Pix, fine-tuned using MagicBrush, the synthetic images are unrealistic and they may be corrupted. However, to the best of our knowledge, there is no traditional technique that would allow us to automatically add objects into an image without altering its main content since the added object may cover some important existing objects in the image. 
In practice, we observe that image-editing deep neural networks try to keep the main portion of the input image unharmed. Thus, we only need to apply some automatic filters to get rid of poor quality images rather than employing a tedious and unscalable manual process. From an input image and a caption, it only takes an average of 9 seconds to generate the corresponding image with clauses and 3 second to assess it by two-stage filtering strategy on an A100 GPU. In total, we spend approximately 86 days to create NeIn with a single A100 GPU.

Fourth, our negative clauses $\mathcal{T}_n$ are descriptive prompts, raising another concerns about whether evaluating image editing models with descriptive prompts is fair. However, existing evidence suggests that image editing methods can effectively handle descriptive prompts.
For instance, “in a race car video game" and “it is now midnight" in from InstructPix2Pix paper; “the dog is looking forward” and “she is now cutting up carrots” from MagicBrush dataset; and “the tarantula is given a glowing outline and the background is changed to a dramatic sunset with vibrant reds and purples” from HQ-Edit dataset. This demonstrates that image editing models must understand descriptive prompts, therefore, they must also understand negative cues, e.g. “the tree is \textit{without} a candle."

\section{Experiments on Image Editing Task}
\label{sec:exp}
\subsection{Experiment Setup}
We benchmark our evaluation set in five SOTA image editing methods, including InstructPix2Pix \cite{instructpix2pix}, MagicBrush \cite{magicbrush}, ZONE \cite{zone}, HQ-Edit \cite{hqedit}, and InstructDiffusion \cite{instructdiffusion}. We fine-tune InstructPix2Pix and MagicBrush in our training set with 8 epochs.
All baseline details, including the versions of BLIP and LLaVA-NeXT, are provided in the supplementary material.

\begin{figure}[h]
\begin{center}
\includegraphics[width=\linewidth]{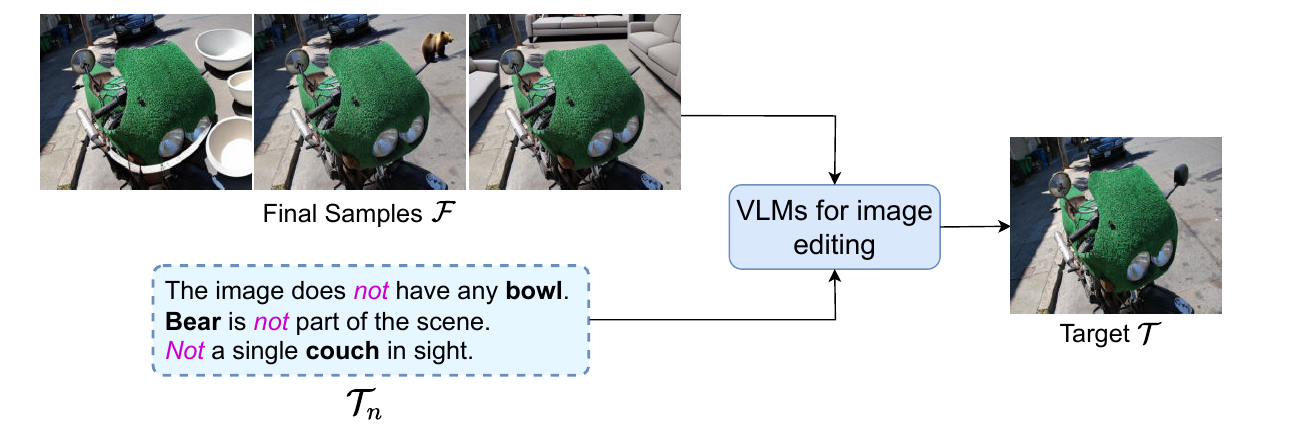}
\caption{Illustration for fine-tuning and benchmarking process.}
\label{fig:finetune}
\vspace{-7mm}
\end{center}
\end{figure}

For each tuple $i^{th}$, these models take $\mathcal{F}^{(i)}$ as an input image and $\mathcal{T}_n^{(i)}$ as an instruction, and return the target image $\mathcal{T}^{(i)}$; where the original image $\mathcal{I}^{(i)}$ is the ground truth; this is illustrated in Figure \ref{fig:finetune}. In fairness, we conduct experiments with the default settings for each model.

\subsection{Evaluation Metrics}

To assess the difference between the output of the image editing models $\mathcal{T}$ and ground truth $\mathcal{I}$, we consider two aspects: image quality and negative instruction satisfaction.

For image quality, we follow MagicBrush \cite{magicbrush}, employing four different metrics: L1, L2, CLIP-I, and DINO. The L1 and L2 metrics measure the pixel-level distance between $\mathcal{T}$ and $\mathcal{I}$. CLIP-I metric leverages CLIP \cite{clip} model, while DINO metric employs DINO \cite{dino} model to compute the alignment between target and query images by measuring the cosine similarity of their embeddings. We consider the image discrepancy in terms of realism using Clean-FID \cite{cleanfid} metric, an improved version of the Frechet Inception Distance (FID) metric \cite{fid} that focuses on image resizing and quantization. In addition, human perceptual judgment is measured by LPIPS \cite{lpips} score to more comprehensively evaluate $\mathcal{T}$ and $\mathcal{I}$ by considering human visual perception.

Then, to measure how much the output semantically satisfies the instruction, we consider whether image editing methods successfully eliminate the object categories specified in the negative sentence, and determine if these methods can preserve the objects not mentioned in the negative sentence.
The first is determined by the \textit{Removal score}, while the second is assessed using the \textit{Retention score}. 

Since the purpose of both metrics is to identify objects, we consider the visual question answering (VQA), represented by LLaVA-NeXT and open-vocabulary object detection (OVD), represented by OWLv2 \cite{minderer2024scaling}. Note that, different from the generation step, we do not accept false positive results for the evaluation metrics. Therefore, both LLaVA-NeXT and OWLv2 are suitable choices.

\begin{algorithm}[!h]
\textbf{Input:} \\
$\mathcal{T}$: considered model's outputs   \\
$\mathcal{S}$: objects to be removed  \\
\textbf{Output:} \\ $s$: removal score
\caption{Removal Evaluation by VQA}
\label{alg:removed}
\begin{algorithmic}[1]
\State $s \coloneqq 0$
\For{each tuple $(\mathcal{T}^{(i)}, \mathcal{S}^{(i)})$ in $(\mathcal{T}, \mathcal{S})$} 
    \Statex \hspace{0.45cm} \textcolor{teal}{\# pre-defined prompt}
    \State $p \gets$ “Does this image contain a/an $\mathcal{S}^{(i)}$?"
    \If{VQA($\mathcal{T}^{(i)}$, $p$) = “No"} \Comment{Object is removed}
        \State $s \gets s+1$
    \EndIf
\EndFor
\State $s \gets s / |\mathcal{T}|$
\State \Return $s$

\end{algorithmic}
\end{algorithm}

\begin{algorithm}[!h]
\textbf{Input:} \\
$\mathcal{F}$: samples of NeIn \\
$\mathcal{T}_o$: original caption from MS-COCO \\
$\mathcal{T}$: considered model's outputs\\
\textbf{Output:} \\ $s$: retention score
\caption{Retention Evaluation by VQA}
\label{alg:retain}
\begin{algorithmic}[1]
\State $s \coloneqq 0$
\For {each tuple ($\mathcal{F}^{(i)}$, $\mathcal{T}_o^{(i)}$, $\mathcal{T}^{(i)}$) in ($\mathcal{F}$, $\mathcal{T}_o$, $\mathcal{T}$)}
    \State $\text{list}^1$ $\coloneqq $[], $\text{list}^2$ $\coloneqq $[]
    \State $\mathcal{O} \gets$ extractor($\mathcal{T}_o^{(i)}$)  
    \Comment{Original objects in $\mathcal{I}$}
    \Statex \hspace{0.45cm} \textcolor{teal}{\# check $\mathcal{O}$ in $\mathcal{F}$}
    \For{each $object$ in $\mathcal{O}$}
        \State $p \gets$ “Does this image contain a/an $object$ ?"
        \State $b \gets$ VQA($\mathcal{F}^{(i)}$, $p$) \Comment{Boolean result}
        \If{$b =$ “Yes"} \Comment{Object is still in $\mathcal{F}^{(i)}$}
            \State append $object$ to $\text{list}^1$
        \EndIf
    \EndFor
    \Statex \hspace{0.45cm} \textcolor{teal}{\# check $\mathcal{O}$ in both $\mathcal{F}$ and $\mathcal{T}$}
    \For{each $object$ in $\text{list}^1$}
        \State $p \gets$ “Does this image contain a/an $object$ ?"
        \State $b \gets$ VQA($\mathcal{T}^{(i)}$, $p$)) \Comment{Boolean result}
        \If{$b =$ “Yes"} \Comment{Object is in  $\mathcal{F}^{(i)}$ \&  $\mathcal{T}^{(i)}$}
            \State append $object$ to $\text{list}^2$
        \EndIf
    \EndFor

    \State $score \gets \text{length of } \text{list}^2 \,/\, \text{length of } \text{list}^1$ 
    \State $s \gets s + score$
\EndFor

\State $s \gets s \,/\, |\mathcal{T}|$
\State \Return $s$
\vspace{0.9mm}
\end{algorithmic}
\end{algorithm}

\begin{table*}[!h]
\centering
\resizebox{\linewidth}{!}{
\begin{tabular}{l|c|c|c|c|c|c|c|c|c|c|c}
\toprule
\multirow{3}{*}{\textbf{Methods}} & \multicolumn{6}{c|}{\textbf{Image Quality}} & \multicolumn{5}{c}{\textbf{Negation Understanding}} \\
\cline{2-12}
& \multirow{2}{*}{L1 $\downarrow$} & \multirow{2}{*}{L2 $\downarrow$} & \multirow{2}{*}{CLIP-I $\uparrow$} & \multirow{2}{*}{DINO $\uparrow$} & \multirow{2}{*}{FID $\downarrow$} & \multirow{2}{*}{LPIPS $\downarrow$} & \multicolumn{2}{c|}{\textbf{LLaVA-NeXT}} & \multicolumn{3}{c}{\textbf{OWLv2}} \\ 
\cline{8-12}
& & & & & & & Removal $\uparrow$ & Retention $\uparrow$ & Removal $\uparrow$ & AUC-Removal $\uparrow$ & Retention $\uparrow$ \\
\hline \hline
InstructPix2Pix \cite{instructpix2pix} & 11.24  & 3.59 & 81.68 & 73.53 & 10.60 & 0.43 & 3.83 & 81.96 & 6.70 & 50.11 & 81.63\\
\rowcolor{lightorange}\textbf{InstructPix2Pix} & \textbf{8.32} & \textbf{2.32} & \textbf{93.11} & \textbf{91.67} & \textbf{4.08} & \textbf{0.33} & \textbf{93.62} & \textbf{98.26} & \textbf{92.66} & \textbf{97.89} & \textbf{95.83}  \\
\hline
MagicBrush \cite{magicbrush} & 8.95 & 2.69 & 88.29 & 84.91 & 7.80 & 0.36 & 5.06 & 93.86 & 8.13 & 52.48 & 91.39\\
\rowcolor{lightorange}\textbf{MagicBrush} & \textbf{8.38} & \textbf{2.35} & \textbf{93.04} & \textbf{91.53} & \textbf{4.15} & \textbf{0.33} & \textbf{92.18} & \textbf{98.21} & \textbf{91.24} & \textbf{97.34} & \textbf{98.07}\\
\hline
ZONE \cite{zone} & 11.95 & 3.67 & 74.12 & 63.18 & 14.95 & 0.46 & 2.93 & 72.38 & 6.47 & 46.04 & 69.07\\
\hline
HQ-Edit \cite{hqedit} & 23.48 & 9.61 & 62.84 & 46.60 & 27.61 & 0.67 & 32.23 & 54.75 & 40.42 & 70.29 & 57.43\\
\hline
InstructDiffusion \cite{instructdiffusion} & 8.54 & 2.54 & 90.57 & 88.62 & 6.89 & 0.34 & 31.46 & 97.55 & 30.00 & 67.99 & 97.58\\
\bottomrule
\end{tabular}}
\caption{Quantitative results of five image editing SOTA methods on the evaluation set of NeIn. All the metrics are in (\%). The InstructPix2Pix and MagicBrush finetuned on NeIn's training set are \hlc[lightorange]{highlighted}. The FID used here is Clean-FID \cite{cleanfid}.}
\label{tab:benchmark}
\end{table*}

\textbf{Removal Evaluation. } The removal evaluation by VQA is illustrated by Algorithm \ref{alg:removed}, while the evaluation using OVD is in the supplementary. For VQA (LLaVA-NeXT), we prompt “Does this image \textit{contain} a/an \{$\mathcal{S}^{(i)}$\}?" If the result is “No," the object is considered successfully removed. For OVD (OWLv2), we give the model output $\mathcal{T}^{(i)}$ and object $\mathcal{S}^{(i)}$.
The result is a list of bounding boxes with confidence scores. 
The object is considered removed if the length of this list is zero.
Addressing the concern that the bounding box may be misclassified, we calculate the Area Under the Curve (AUC) regarding the highest confidence score of each sample.

\textbf{Retention Evaluation. }We observe that, in the case of not being able to understand which object needs to be removed, the model may still achieve a high removal score by removing as many objects as possible from the images. Hence, we measure the retention score that assess whether or not the model retains the \textit{salient} objects in the original images. Algorithm \ref{alg:retain} shows the retention evaluation by VQA, with the corresponding evaluation for OVD provided in the supplementary.
Let's denote the original objects in $\mathcal{T}_o$ (i.e. objects have in source image $\mathcal{I}$) are $\mathcal{O}$. To split $\mathcal{O}$ from $\mathcal{T}_o$, we use Spacy \cite{spacy}, a popular library for Natural Language Processing. Note that we only consider the \textit{main} objects, which are essential to the content of $\mathcal{I}$, therefore we use $\mathcal{T}_o$ because it covers the important objects in $\mathcal{I}$.

We first check that $\mathcal{O}$ is still present in the samples $\mathcal{F}$ of NeIn. If $\mathcal{O}$ exists in $\mathcal{F}$, we then check whether or not $\mathcal{O}$ is present in generated image $\mathcal{T}$ of considered model.
We divide the number of retained objects by the number of $\mathcal{O}$ to get the retention score for each sample. The final retention score is the average across all samples of the evaluation set.

\begin{figure*}[t]
\begin{center}
\includegraphics[width=\linewidth]{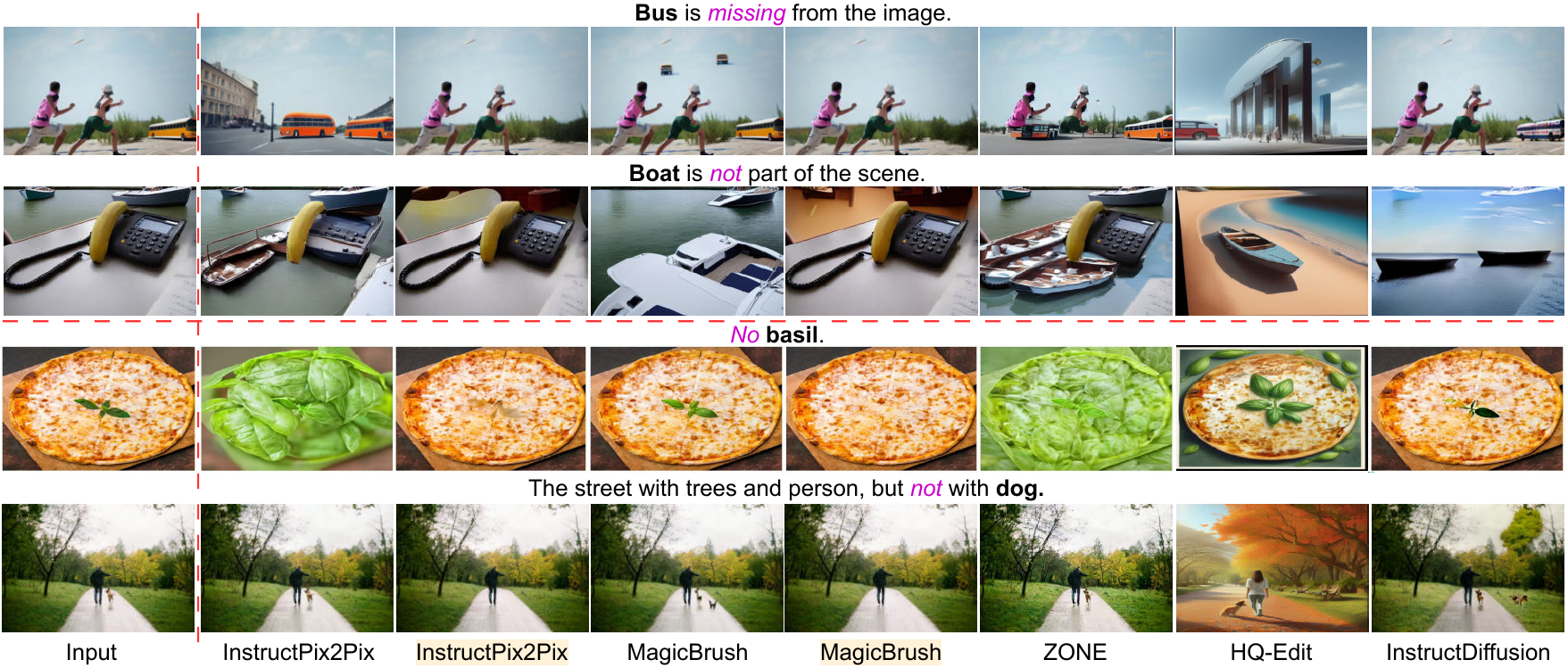}
\caption{Qualitative results of five SOTA methods on NeIn’s evaluation samples (first two samples) and random image-prompt pairs (last two samples). The fine-tuned InstructPix2Pix ($3^{\text{rd}}$  column) and MagicBrush ($5^{\text{th}}$  column) on NeIn's training set are \hlc[lightorange]{highlighted}.}
\label{fig:benchmark}
\end{center}
\vspace{-6mm}
\end{figure*}

\subsection{Results}
\textbf{Quantitative Evaluation. }Table \ref{tab:benchmark} presents the quantitative results between InstructPix2Pix, fine-tuned InstructPix2Pix, MagicBrush, fine-tuned MagicBrush, ZONE, HQ-Edit, and InstructDiffusion on the evaluation set of NeIn.

None of the five methods perform well on pixel-level metrics, such as L1 and L2, or on image quality metrics, such as CLIP-I, DINO, FID, and LPIPS, indicating that negative prompts are considerably challenging. This is particularly evident when considering the Removal and Retention scores. Image editing models generally struggle to understand the meaning of negation when they do not remove the mentioned objects, as demonstrated by the low Removal, calculated by LLaVA-NeXT and OWLv2, and AUC-Removal scores of OWLv2. InstructPix2Pix \cite{instructpix2pix}, ZONE \cite{zone}, and HQ-Edit \cite{hqedit} also distort the content of the source image, as can be seen by the low Retention score. It is noteworthy that HQ-Edit achieves better Removal score and AUC-Removal than the others but worse in retaining the original content of the images.  We hypothesize that the model dramatically alters the images that makes their content indecipherable, as indicated by the high L1 and L2 scores. InstructDiffusion \cite{instructdiffusion} handles negation best among the five baselines, likely due to its training on diverse computer vision datasets that enhance generalization for negative queries.

In contrast, the fine-tuned versions of InstructPix2Pix and MagicBrush significantly enhance both image quality and instruction satisfaction with negative queries. 
Given that current image editing methods rely on Diffusion \cite{stablediffusion} framework. We assume by this evidence, other text-guided image editing methods are likely to achieve similar results.

\textbf{Qualitative Evaluation. }
Some results for each image editing model are illustrated in Figure \ref{fig:benchmark}. 
We consider both image-prompt pairs from NeIn's samples and randomly pairs outside of NeIn's distribution to evaluate the generalization of fine-tuned versions.

Instead of removing the mentioned objects, original image editing models tend to have the following problems: (1) retaining the mentioned object in the edited image; (2) increasing the quantity of mentioned object in the generated image, and even bringing that object to the center of the images; and (3) completely replacing the content of the query image with that object. This observation demonstrates the failure of VLMs in image editing on negation understanding, potentially affecting other vision-language tasks. 

On the contrary, the fine-tuned InstructPix2Pix and MagicBrush models clearly demonstrate the ability to remove objects specified in negative queries. 
Even when dealing with difficult prompt that include trees, person, and dog; models are still able to successfully understand negation.
This suggests that, following fine-tuning with NeIn, VLMs may be capable of understanding negation.
\section{Conclusion}
We introduce NeIn, the first large-scale dataset for negation understanding within the context of text-guided image editing, comprising 366,957 quintuplets with 342,775 queries for training set and 24,182 queries for evaluation set. Negation understanding, an important linguistic concept yet to be fully explored in image editing, is a crucial task for aligning image editing VLMs with human information needs. We present a novel pipeline to automatically generate and filter samples for NeIn by leveraging VLMs for vision-language pre-training and image editing. Additionally, we present a comprehensive evaluation protocol, including removal and retention aspects, to assess the performance of current image editing models on negation understanding for NeIn's evaluation set. Our experiments reveal that existing image editing methods struggle to understand negative queries, highlighting a new challenge for the research community. By fine-tuning these models on NeIn's training set, we can improve their ability to identify negative terms in user queries, as demonstrated by both quantitative and qualitative results.

\textbf{Limitations. }Two current limitations of NeIn are that (1) we have only performed experiments using image editing models, and (2) the negative predefined prompts are relatively simple. \textbf{Future Directions. } Based on current limitations, future research plans to leverage and expand NeIn include: (1) fine-tuning and benchmarking NeIn for other tasks in vision-language domain such as composed image retrieval and image-text matching; (2) considering complex negative sentences involving words such as “except", “neither-nor", etc. We hope that with the release of NeIn, the research community will shift its focus toward negation understanding, which we believe is an important open problem for VLMs.

{
    \small
    \bibliographystyle{ieeenat_fullname}
    \bibliography{main}

\begin{thebibliography}{30}
\providecommand{\natexlab}[1]{#1}
\providecommand{\url}[1]{\texttt{#1}}
\expandafter\ifx\csname urlstyle\endcsname\relax
  \providecommand{\doi}[1]{doi: #1}\else
  \providecommand{\doi}{doi: \begingroup \urlstyle{rm}\Url}\fi

\bibitem[Brooks et~al.(2023)Brooks, Holynski, and Efros]{instructpix2pix}
Tim Brooks, Aleksander Holynski, and Alexei~A Efros.
\newblock Instructpix2pix: Learning to follow image editing instructions.
\newblock In \emph{CVPR}, 2023.

\bibitem[Caron et~al.(2021)Caron, Touvron, Misra, J{\'e}gou, Mairal, Bojanowski, and Joulin]{dino}
Mathilde Caron, Hugo Touvron, Ishan Misra, Herv{\'e} J{\'e}gou, Julien Mairal, Piotr Bojanowski, and Armand Joulin.
\newblock Emerging properties in self-supervised vision transformers.
\newblock In \emph{ICCV}, 2021.

\bibitem[Changpinyo et~al.(2021)Changpinyo, Sharma, Ding, and Soricut]{cc12m}
Soravit Changpinyo, Piyush Sharma, Nan Ding, and Radu Soricut.
\newblock Conceptual 12m: Pushing web-scale image-text pre-training to recognize long-tail visual concepts.
\newblock In \emph{CVPR}, 2021.

\bibitem[Geng et~al.(2024)Geng, Yang, Hang, Li, Gu, Zhang, Bao, Zhang, Li, Hu, et~al.]{instructdiffusion}
Zigang Geng, Binxin Yang, Tiankai Hang, Chen Li, Shuyang Gu, Ting Zhang, Jianmin Bao, Zheng Zhang, Houqiang Li, Han Hu, et~al.
\newblock Instructdiffusion: A generalist modeling interface for vision tasks.
\newblock In \emph{CVPR}, 2024.

\bibitem[Heusel et~al.(2017)Heusel, Ramsauer, Unterthiner, Nessler, and Hochreiter]{fid}
Martin Heusel, Hubert Ramsauer, Thomas Unterthiner, Bernhard Nessler, and Sepp Hochreiter.
\newblock Gans trained by a two time-scale update rule converge to a local nash equilibrium.
\newblock \emph{NeurIPS}, 2017.

\bibitem[Honnibal et~al.(2020)Honnibal, Montani, Van~Landeghem, Boyd, et~al.]{spacy}
Matthew Honnibal, Ines Montani, Sofie Van~Landeghem, Adriane Boyd, et~al.
\newblock spacy: Industrial-strength natural language processing in python.
\newblock 2020.

\bibitem[Horn(2010)]{negation}
Laurence~R. Horn, editor.
\newblock \emph{The Expression of Negation}.
\newblock De Gruyter Mouton, 2010.

\bibitem[Hui et~al.(2024)Hui, Yang, Zhao, Shi, Wang, Wang, Zhou, and Xie]{hqedit}
Mude Hui, Siwei Yang, Bingchen Zhao, Yichun Shi, Heng Wang, Peng Wang, Yuyin Zhou, and Cihang Xie.
\newblock Hq-edit: A high-quality dataset for instruction-based image editing.
\newblock \emph{arXiv preprint arXiv:2404.09990}, 2024.

\bibitem[Li et~al.(2022)Li, Li, Xiong, and Hoi]{blip}
Junnan Li, Dongxu Li, Caiming Xiong, and Steven Hoi.
\newblock Blip: Bootstrapping language-image pre-training for unified vision-language understanding and generation.
\newblock In \emph{ICML}, 2022.

\bibitem[Li et~al.(2024)Li, Zeng, Feng, Gao, Liu, Liu, Li, Tang, Hu, Liu, et~al.]{zone}
Shanglin Li, Bohan Zeng, Yutang Feng, Sicheng Gao, Xiuhui Liu, Jiaming Liu, Lin Li, Xu Tang, Yao Hu, Jianzhuang Liu, et~al.
\newblock Zone: Zero-shot instruction-guided local editing.
\newblock In \emph{CVPR}, 2024.

\bibitem[Lin et~al.(2014)Lin, Maire, Belongie, Hays, Perona, Ramanan, Doll{\'a}r, and Zitnick]{coco}
Tsung-Yi Lin, Michael Maire, Serge Belongie, James Hays, Pietro Perona, Deva Ramanan, Piotr Doll{\'a}r, and C~Lawrence Zitnick.
\newblock Microsoft coco: Common objects in context.
\newblock In \emph{ECCV}, 2014.

\bibitem[Liu et~al.(2024)Liu, Li, Li, Li, Zhang, Shen, and Lee]{llavanext}
Haotian Liu, Chunyuan Li, Yuheng Li, Bo Li, Yuanhan Zhang, Sheng Shen, and Yong~Jae Lee.
\newblock Llava-next: Improved reasoning, ocr, and world knowledge, 2024.

\bibitem[Liu et~al.(2021)Liu, Rodriguez-Opazo, Teney, and Gould]{cirr}
Zheyuan Liu, Cristian Rodriguez-Opazo, Damien Teney, and Stephen Gould.
\newblock Image retrieval on real-life images with pre-trained vision-and-language models.
\newblock In \emph{ICCV}, 2021.

\bibitem[Mai et~al.(2024)Mai, Gauch, and Adams]{mai2024bert}
Quan Mai, Susan Gauch, and Douglas Adams.
\newblock Setbert: Enhancing retrieval performance for boolean logic and set operation queries.
\newblock \emph{arXiv preprint arXiv:2406.17282}, 2024.

\bibitem[Mai et~al.(2025)Mai, Gauch, and Adams]{mai2025boolean}
Quan Mai, Susan Gauch, and Douglas Adams.
\newblock Boolean-aware attention for dense retrieval.
\newblock \emph{arXiv preprint arXiv:2503.01753}, 2025.

\bibitem[Minderer et~al.(2024)Minderer, Gritsenko, and Houlsby]{minderer2024scaling}
Matthias Minderer, Alexey Gritsenko, and Neil Houlsby.
\newblock Scaling open-vocabulary object detection.
\newblock \emph{NeurIPS}, 2024.

\bibitem[Ordonez et~al.(2011)Ordonez, Kulkarni, and Berg]{sbu}
Vicente Ordonez, Girish Kulkarni, and Tamara Berg.
\newblock Im2text: Describing images using 1 million captioned photographs.
\newblock \emph{NeurIPS}, 2011.

\bibitem[Parmar et~al.(2022)Parmar, Zhang, and Zhu]{cleanfid}
Gaurav Parmar, Richard Zhang, and Jun-Yan Zhu.
\newblock On aliased resizing and surprising subtleties in gan evaluation.
\newblock In \emph{CVPR}, 2022.

\bibitem[Radford et~al.(2021)Radford, Kim, Hallacy, Ramesh, Goh, Agarwal, Sastry, Askell, Mishkin, Clark, et~al.]{clip}
Alec Radford, Jong~Wook Kim, Chris Hallacy, Aditya Ramesh, Gabriel Goh, Sandhini Agarwal, Girish Sastry, Amanda Askell, Pamela Mishkin, Jack Clark, et~al.
\newblock Learning transferable visual models from natural language supervision.
\newblock In \emph{ICML}, 2021.

\bibitem[Ravichander et~al.(2022)Ravichander, Gardner, and Marasovi{\'c}]{ravichander2022condaqa}
Abhilasha Ravichander, Matt Gardner, and Ana Marasovi{\'c}.
\newblock Condaqa: A contrastive reading comprehension dataset for reasoning about negation.
\newblock \emph{arXiv preprint arXiv:2211.00295}, 2022.

\bibitem[Rombach et~al.(2022)Rombach, Blattmann, Lorenz, Esser, and Ommer]{stablediffusion}
Robin Rombach, Andreas Blattmann, Dominik Lorenz, Patrick Esser, and Bj{\"o}rn Ommer.
\newblock High-resolution image synthesis with latent diffusion models.
\newblock In \emph{CVPR}, 2022.

\bibitem[Schuhmann et~al.(2021)Schuhmann, Vencu, Beaumont, Kaczmarczyk, Mullis, Katta, Coombes, Jitsev, and Komatsuzaki]{laion}
Christoph Schuhmann, Richard Vencu, Romain Beaumont, Robert Kaczmarczyk, Clayton Mullis, Aarush Katta, Theo Coombes, Jenia Jitsev, and Aran Komatsuzaki.
\newblock Laion-400m: Open dataset of clip-filtered 400 million image-text pairs.
\newblock \emph{arXiv preprint arXiv:2111.02114}, 2021.

\bibitem[Sharma et~al.(2018)Sharma, Ding, Goodman, and Soricut]{cc3m}
Piyush Sharma, Nan Ding, Sebastian Goodman, and Radu Soricut.
\newblock Conceptual captions: A cleaned, hypernymed, image alt-text dataset for automatic image captioning.
\newblock In \emph{Proceedings of the 56th Annual Meeting of the Association for Computational Linguistics (Volume 1: Long Papers)}, 2018.

\bibitem[Singh et~al.(2024)Singh, Shrivastava, Vatsa, Singh, and Bharati]{singh2024learn}
Jaisidh Singh, Ishaan Shrivastava, Mayank Vatsa, Richa Singh, and Aparna Bharati.
\newblock Learn" no" to say" yes" better: Improving vision-language models via negations.
\newblock \emph{arXiv preprint arXiv:2403.20312}, 2024.

\bibitem[Truong et~al.(2023)Truong, Baldwin, Verspoor, and Cohn]{truong2023language}
Thinh~Hung Truong, Timothy Baldwin, Karin Verspoor, and Trevor Cohn.
\newblock Language models are not naysayers: an analysis of language models on negation benchmarks.
\newblock \emph{arXiv preprint arXiv:2306.08189}, 2023.

\bibitem[Wang et~al.(2022)Wang, Chen, Hu, and Li]{wang2022learn}
Ziyue Wang, Aozhu Chen, Fan Hu, and Xirong Li.
\newblock Learn to understand negation in video retrieval.
\newblock In \emph{ACMMM}, 2022.

\bibitem[Weller et~al.(2023)Weller, Lawrie, and Van~Durme]{weller2023nevir}
Orion Weller, Dawn Lawrie, and Benjamin Van~Durme.
\newblock Nevir: negation in neural information retrieval.
\newblock \emph{arXiv preprint arXiv:2305.07614}, 2023.

\bibitem[Zhang et~al.(2024{\natexlab{a}})Zhang, Mo, Chen, Sun, and Su]{magicbrush}
Kai Zhang, Lingbo Mo, Wenhu Chen, Huan Sun, and Yu Su.
\newblock Magicbrush: A manually annotated dataset for instruction-guided image editing.
\newblock \emph{NeurIPS}, 2024{\natexlab{a}}.

\bibitem[Zhang et~al.(2018)Zhang, Isola, Efros, Shechtman, and Wang]{lpips}
Richard Zhang, Phillip Isola, Alexei~A Efros, Eli Shechtman, and Oliver Wang.
\newblock The unreasonable effectiveness of deep features as a perceptual metric.
\newblock In \emph{CVPR}, 2018.

\bibitem[Zhang et~al.(2024{\natexlab{b}})Zhang, Zhang, Wu, Pei, Ren, de~Rijke, Chen, and Ren]{zhang2024excluir}
Wenhao Zhang, Mengqi Zhang, Shiguang Wu, Jiahuan Pei, Zhaochun Ren, Maarten de Rijke, Zhumin Chen, and Pengjie Ren.
\newblock Excluir: Exclusionary neural information retrieval.
\newblock \emph{arXiv preprint arXiv:2404.17288}, 2024{\natexlab{b}}.

\end{thebibliography}
}
\thispagestyle{empty}
\appendix


\clearpage

\begin{figure*}[h]
\begin{center}
\includegraphics[width=\linewidth]{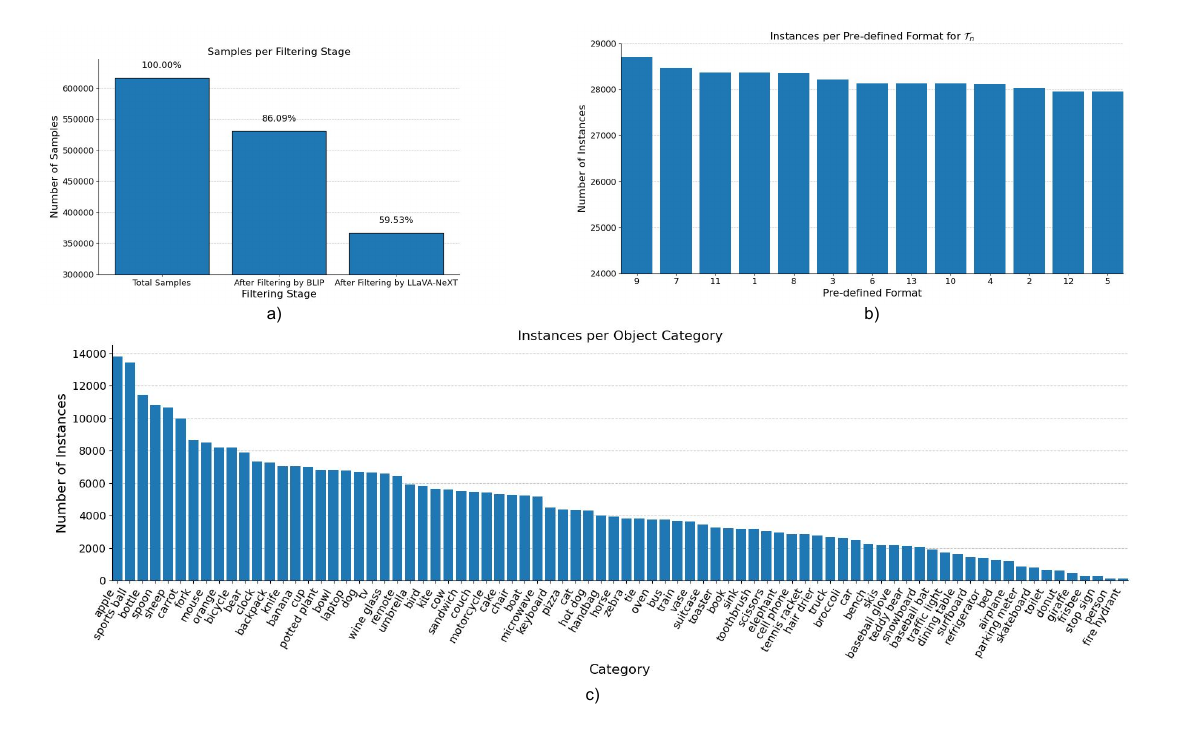}
\vspace{-5mm}
\caption{Statistical analysis for NeIn. a) Number of samples after each filtering phase; b) Number of instances per pre-defined format for $\mathcal{T}_{n}$, the x-axis is followed to the order presented Table 2; c) Number of instances per object category, these 80 objects follow the objects in MS-COCO dataset. Best view in zoom.}
\vspace{-3mm}
\label{fig:chart}
\end{center}
\end{figure*}

\section{Dataset Statistics}
The MS-COCO \cite{coco} dataset includes a total of 123,287 images with 616,435 captions for training and evaluation. For each COCO caption, we generate a synthetic image for NeIn, resulting in 616,435 synthetic images in total. 
Following the first filtering stage using  BLIP \cite{blip}, NeIn contains a total of 530,694 samples, accounting for 86.1\% of the dataset and excluding 85,741 (13.9\%) erroneous entries. 
Subsequently, after filtering with LLaVA-NeXT \cite{llavanext}, NeIn retains 366,957 samples (59.53\% of the original set). In total, the two-stage filtering process eliminates 40.47\% of erroneous samples. Figure \ref{fig:chart}(a) illustrates the number of samples after each filtering phase.

In the interest of efficiency, we randomly select 24,182 queries for benchmarking, while the remaining 342,775 queries are used for fine-tuning. This data split is suitable for image editing task, with approximately 6.6\% for the evaluation set, since the evaluation set of MagicBrush accounts for 6\%.

In order to exam the distribution within finalized dataset, i.e., after filtering process, we present the distribution of the pre-defined format for $\mathcal{T}_{n}$, as depicted in Figure \ref{fig:chart}(b). 

We also analyze the number of instances per category for 80 object categories in Figure \ref{fig:chart}(c).
Since NeIn is designed to address the problem of negation understanding, object imbalance does not affect the quality of our dataset. This imbalance phenomenon may be due to the difficulty of incorporating objects with low proportion in the distribution (e.g., stop sign, frisbee) into the context of COCO images using MagicBrush \cite{magicbrush}. Consequently, generated images containing these objects are often filtered out by BLIP or LLaVA-NeXT. 

\section{Dataset Format}
Table \ref{tab:format} provides the format for one sample in the JSON file. Each sample in NeIn consists of five components: the source image, original caption, generated sentence, negative sentence, and the NeIn sample itself. These components are represented in the JSON file as “COCO" ($\mathcal{I}$), “T\_original" ($\mathcal{T}_o$), “T\_generated" ($\mathcal{T}_g$), “T\_negative" ($\mathcal{T}_n$), and “NeIn" ($\mathcal{F}$), respectively.

For each caption in COCO, we can generate a corresponding sample for NeIn, the “NeIn\_000000000074\_5" indicates that this image is derived from the COCO image “COCO\_000000000074,” utilizing its fifth caption.

Based on the “NeIn" sample and the “T\_negative" clause, the ground truth corresponds to the “COCO" image. The “T\_original" from COCO is employed for the \textit{retention} evaluation, whereas the “T\_generated" is utilized to extract the selected object category ($\mathcal{S}$) for the \textit{removal} evaluation.

\begin{table}[!h]
\begin{adjustbox}{max width=\linewidth}
\begin{tcolorbox}[colback=white, colframe=black, boxrule=0.5pt]
\ttfamily
\textbf{\{}\\
\ \ \ \textcolor{mygreen}{\textbf{"COCO"}}\textbf{:} \textcolor{myred}{"COCO\_000000000074"}\textbf{,}\\
\ \ \ \textcolor{mygreen}{\textbf{"T\_original"}}\textbf{:} \textcolor{myred}{"A puppy rests on the street next to a bicycle."}\textbf{,}\\
\ \ \ \textcolor{mygreen}{\textbf{"T\_generated"}}\textbf{:} \textcolor{myred}{"Add a couch."}\textbf{,}\\
\ \ \ \textcolor{mygreen}{\textbf{"T\_negative"}}\textbf{:} \textcolor{myred}{"The image cannot have any couch."}\textbf{,}\\
\ \ \ \textcolor{mygreen}{\textbf{"NeIn"}}\textbf{:} \textcolor{myred}{"NeIn\_000000000074\_5"}\\
\textbf{\}}
\end{tcolorbox}
\end{adjustbox}
\vspace{-5mm}
\caption{The JSON format of NeIn.} 
\label{tab:format}
\end{table}

\section{NeIn's Sample After Filtering Process}

\begin{figure*}[t]
\begin{center}
\includegraphics[width=\linewidth]{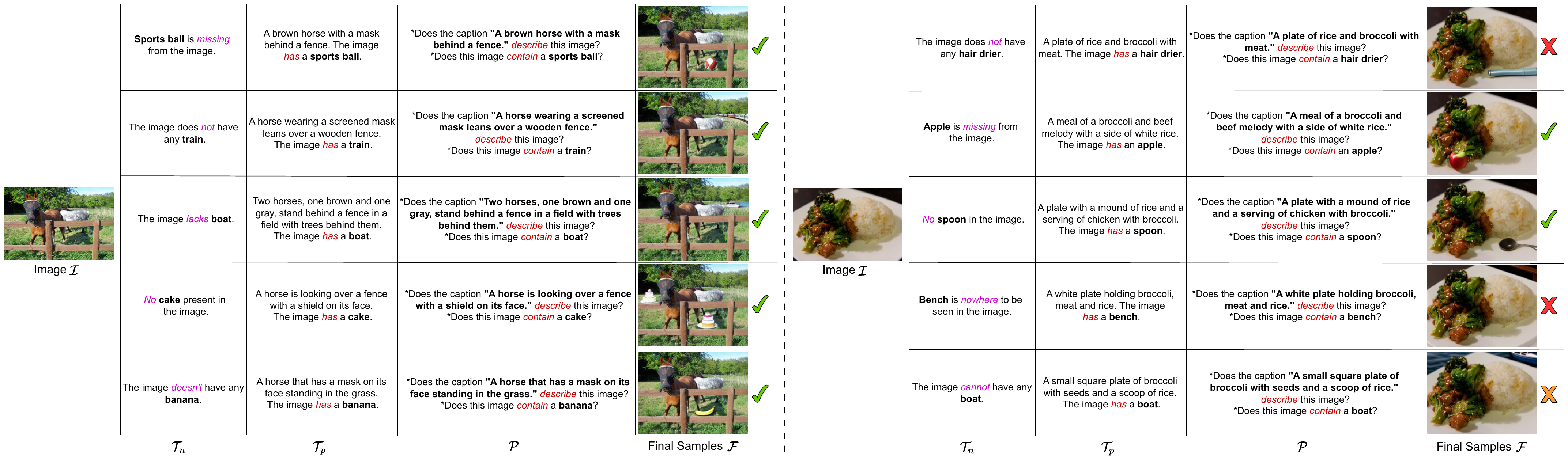}
\caption{Two examples for our filtering process. The \textcolor{green}{green} \cmark \hspace{0.1mm} indicates samples that are retained. The \textcolor{red}{red} \xmark \hspace{0.1mm} signifies samples removed by BLIP, while the \textcolor{orange}{orange} \xmark \hspace{0.1mm} checkmark denotes samples removed by LLaVA-NeXT. Best view in zoom.}
\label{fig:filtering}
\end{center}
\vspace{-2mm}
\end{figure*}

Figure \ref{fig:filtering} illustrates the examples of NeIn after the two-stage filtering process. In the first example, all samples are retained after filtering. This occurs when the final samples $\mathcal{F}$ pass both the BLIP and LLaVA-NeXT checks. In contrast, in the second example, the first, fourth, and fifth samples are rejected. The first and fourth samples are removed because the objects added by MagicBrush are not recognized as a hair dryer and bench by BLIP. In the case of the fifth sample, BLIP recognizes the presence of a boat, whereas LLaVA-NeXT does not assent its presence in the image. Using a two-stage filtering process with image-text matching (BLIP) and visual question answering (LLaVA-NeXT) ensures the high quality for NeIn's samples.


\section{Baseline Details}
We leverage API from Transformers library\footnote{https://huggingface.co/docs/transformers/index} for both BLIP and LLaVA-NeXT. We use the large version for BLIP\footnote{https://huggingface.co/Salesforce/blip-itm-large-coco}. We select BLIP over more recent  vision-language models for image-text matching in generation phase because it is trained on MS-COCO dataset, and the objects (e.g., “car," “apple") are relatively simple compared to complex texts that BLIP can perform. While for the filtering phase, BLIP is additionally supported by LLaVA-NeXT. Leveraging BLIP accelerates the process of generating datasets with a large number of samples. The threshold of 0.4 is selected based on experiments.
For LLaVA-NeXT, we employ the Mistral-7B version\footnote{https://huggingface.co/liuhaotian/llava-v1.6-mistral-7b} to strike a balance between accuracy and resource efficiency.

We evaluate NeIn on five SOTA methods published in 2023 and 2024: InstructPix2Pix \cite{instructpix2pix}\footnote{https://github.com/timothybrooks/instruct-pix2pix}, MagicBrush \cite{magicbrush}\footnote{https://github.com/OSU-NLP-Group/MagicBrush}, ZONE \cite{zone}\footnote{https://github.com/lsl001006/ZONE}, HQ-Edit \cite{hqedit}\footnote{https://github.com/UCSC-VLAA/HQ-Edit}, and InstructDiffusion \cite{instructdiffusion}\footnote{https://github.com/cientgu/InstructDiffusion}:

\begin{algorithm}[!h]
\textbf{Input:} \\
$\mathcal{T}$: considered model's outputs   \\
$\mathcal{S}$: objects to be removed  \\
\textbf{Output:} \\ $s$: removal score
\caption{Removal Evaluation by OVD}
\label{alg:removed_ovd}
\begin{algorithmic}[1]
\State $s \coloneqq 0$
\For{each tuple $(\mathcal{T}^{(i)}, \mathcal{S}^{(i)})$ in $(\mathcal{T}, \mathcal{S})$} 

        \State $p \gets$ OVD($\mathcal{T}^{(i)}$, $\mathcal{S}^{(i)}$) \Comment{Prediction list}
        \If{$\text{length of } p = 0$} \Comment{Object is removed}
            \State $s \gets s+1$
        \EndIf

\EndFor

\State $s \gets s / |\mathcal{T}|$
\State \Return $s$

\end{algorithmic}
\end{algorithm}

\begin{figure*}[!h]
\begin{center}
\includegraphics[width=\linewidth]{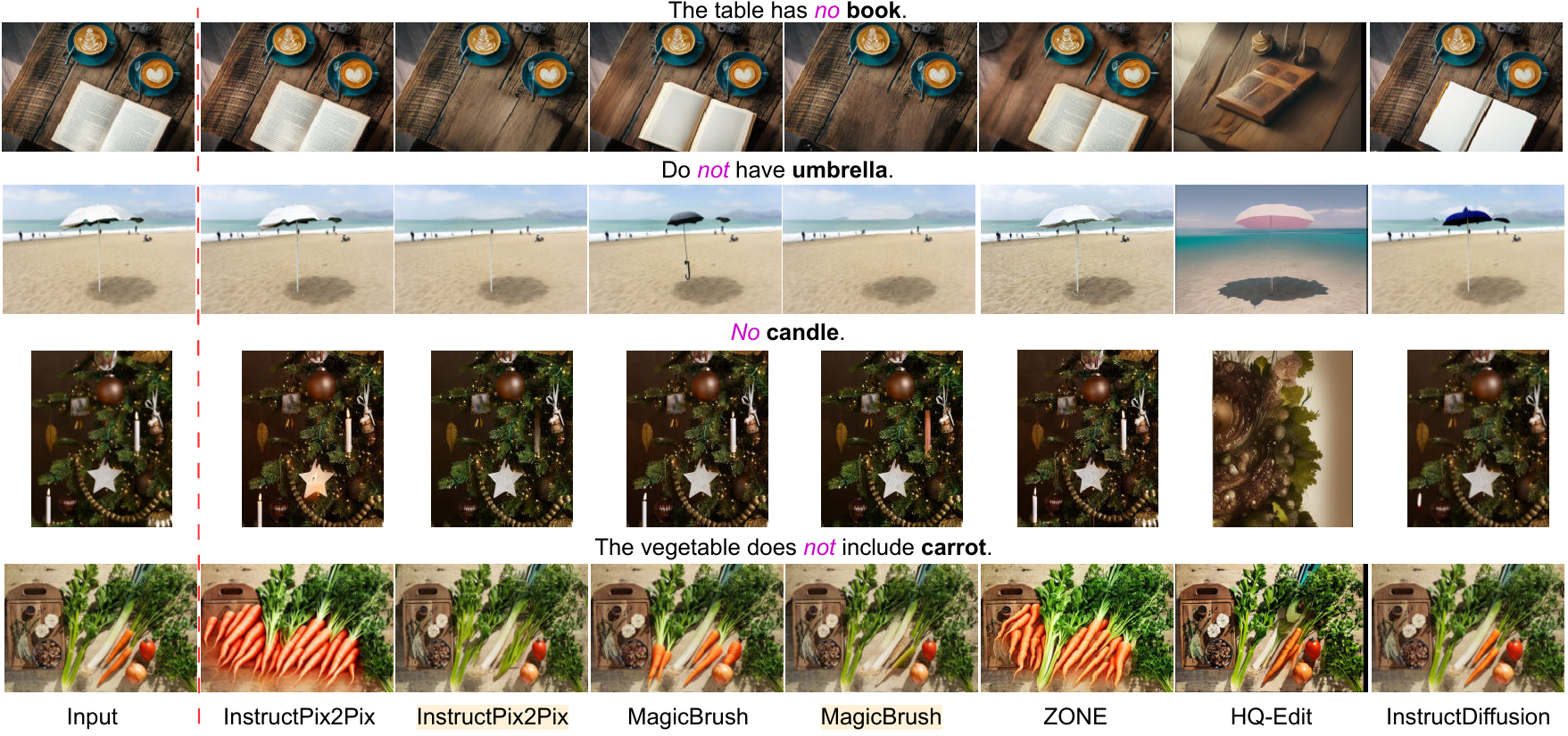}
\caption{Additional qualitative results of five SOTA methods. The fine-tuned InstructPix2Pix ($3^{\text{rd}}$  column) and MagicBrush ($5^{\text{th}}$  column) on NeIn's training set are \hlc[lightorange]{highlighted}.}
\vspace{-7mm}
\label{fig:benchmark_supp}
\end{center}
\end{figure*}

1) InstructPix2Pix proposes a multi-modal training dataset 
comprising input images, text-based editing instructions, and the corresponding edited images. It fine-tunes Stable Diffusion \cite{stablediffusion} by this dataset in a supervised manner to achieve zero-shot image editing.

2) MagicBrush is a manually annotated dataset with 10,388 samples covering both single- and multi-turn image editing scenarios. The fine-tuned version for InstructPix2Pix demonstrates superior editing performance compared to other approaches.

3) ZONE initially utilizes InstructPix2Pix to identify the editing locations given the text instructions. It then refines those regions using the Region-IoU scheme combined with the Fourier-based edge smoother. ZONE produces high-quality results for intuitive instructions such as “add”, “remove”, and “change”.

\begin{figure*}[!h]
\begin{center}
\includegraphics[width=\linewidth]{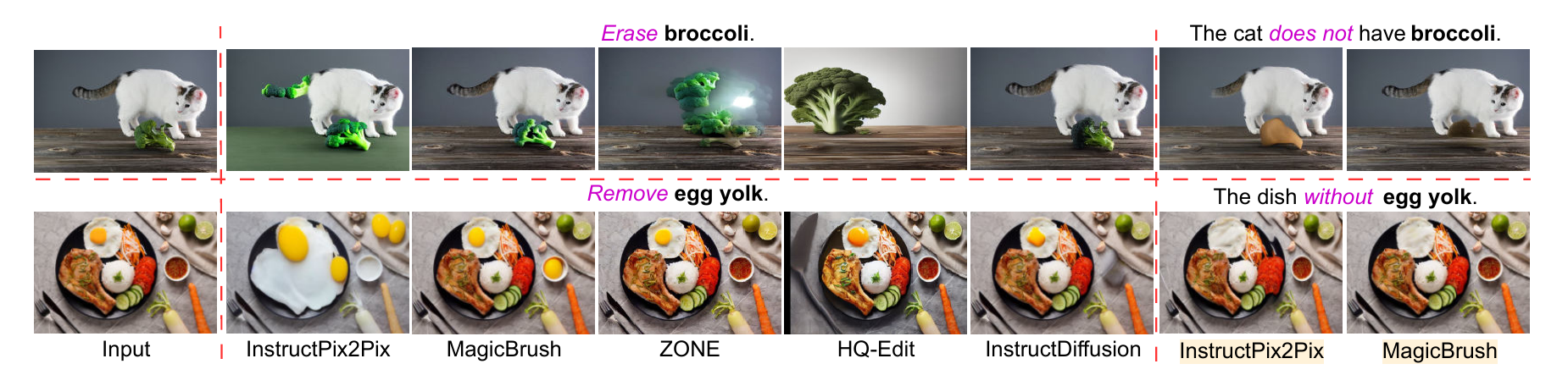}
\caption{Direct and negative instruction. The fine-tuned InstructPix2Pix and MagicBrush on NeIn's training set are \hlc[lightorange]{highlighted} in the last two columns. Even though our fine-tuned InstructPix2Pix can't perfectly remove the broccoli in the first example, it still recognizes that the broccoli should be removed.}
\vspace{-7mm}
\label{fig:benchmark_supp}
\end{center}
\end{figure*}

4) HQ-Edit introduces a new image editing dataset comprising approximately 200,000 edits, by leveraging the large language and text-to-image models. The fine-tuned version for InstructPix2Pix generates high-quality edited results, further validating its effectiveness for image editing.

5) InstructionDiffusion generalizes various computer vision tasks as instruction-based image editing. In order to achieve that goal, it combines multiple datasets for  keypoint detection, semantic segmentation, referring segmentation, image enhancement (e.g., denoising, deblurring, and watermark removal), and image editing. The results indicate that the multi-task learning strategy is benefit for image editing.


\section{Removal and Retention Evaluation by Open-Vocabulary Detection (OVD)}
Algorithm \ref{alg:removed_ovd} and algorithm \ref{alg:retain_ovd} illustrate the removal and retention evaluation by OVD (OWLv2). Instead of using predefined prompts as in VQA, we leverage the capabilities of an open-vocabulary object detection model, specifically OWLv2, to detect arbitrary objects in images.

\section{Additional Qualitative Results}
We provide more qualitative results of five baselines and two fine-tuned versions for negative prompts in Figure \ref{fig:benchmark_supp}. The original versions of the baseline models generally fail to recognize negative words. They tend to either retain the object in the image, replace the object with a similar one of a different appearance, add more objects to the image, or completely alter the content of the query image.

\begin{algorithm}[!h]
\textbf{Input:} \\
$\mathcal{F}$: samples of NeIn \\
$\mathcal{T}_o$: original caption from MS-COCO \\
$\mathcal{T}$: considered model's outputs\\
\textbf{Output:} \\
$s$: retention score
\caption{Retention Evaluation by OVD}
\label{alg:retain_ovd}
\begin{algorithmic}[1]
\State $s \coloneqq 0$
\For {each tuple ($\mathcal{F}^{(i)}$, $\mathcal{T}_o^{(i)}$, $\mathcal{T}^{(i)}$) in ($\mathcal{F}$, $\mathcal{T}_o$, $\mathcal{T}$)}
    \State $\text{list}^1$ $\coloneqq $[], $\text{list}^2$ $\coloneqq $[]
    \State $\mathcal{O} \gets$ extractor($\mathcal{T}_o^{(i)}$)  
    \Comment{Original objects in $\mathcal{I}$}

    \State $p^1 \gets$ OVD($\mathcal{F}^{(i)}$, $\mathcal{O}$)  
    \Comment{Objects are still in $\mathcal{F}^{(i)}$}

    \For{each $object$ in $p^1$}
    \Statex \hspace{0.95cm} \textcolor{teal}{\# each object in $p^1$ may overlap with multiple confidence scores; store each object only once}
        \State append unique $object$ to $\text{list}^1$  
    \EndFor

    \State $p^2 \gets$ OVD($\mathcal{T}^{(i)}$, $\text{list}^1$)
    \Comment{Objects in  $\mathcal{F}^{(i)}$ \&  $\mathcal{T}^{(i)}$}
    \For{each $object$ in $p^2$}
        \State append unique $object$ to $\text{list}^2$
    \EndFor

    \State $score \gets \text{length of } \text{list}^2  \,/\, \text{length of } \text{list}^1$ 
    \State $s \gets s + score$
\EndFor

\State $s \gets s \,/\, |\mathcal{T}|$
\State \Return $s$

\end{algorithmic}
\end{algorithm}

InstructDiffusion occasionally handles negative queries effectively, as demonstrated by the third sample and its relatively high scores in removal and retention metrics. We hypothesize that the combination of datasets from various computer vision tasks enhances the model's generalization capabilities, therefore, improving its understanding of negation. HQ-Edit is capable of removing objects specified in the negative prompts from the query image. However, it often significantly alters the overall content of the query image. This suggests that HQ-Edit may not fully understand the concept of negation; instead, it likely modifies the image content based on the characteristics it learned from its training dataset.

After being fine-tuned on our training set, InstructPix2Pix and MagicBrush are capable of handling negative queries, even with verbs not included in the pre-defined prompts (e.g., “include" in the last query) and nouns absent from the MS-COCO dataset (e.g., “candle" in the third query). We encourage researchers to continue leveraging NeIn for other tasks within the vision-language domain.

\section{Direct Instruction and Negation}

Although the “remove" and “erase"  instructions appear in text-guided image editing datasets (e.g., MagicBrush contains 7\% remove prompts), original image editing models still struggle to remove objects from images. Our fine-tuned models, which leverage negation, provide an alternative yet still natural approach for object removal. Also, in practice, negation is used frequently by human, and we cannot expect users to avoid using negative words with image editing models. We hope that combining NeIn with existing image editing datasets will help models meet user expectations.

\section{Future Directions in Other Vision-Language Tasks}
While designed for text-guided image editing, NeIn is potentially suitable for evaluating negation understanding in other vision-language tasks. 
For distinct tasks, we can use NeIn in appropriate ways to assess models' negation understanding. \textit{For composed image retrieval}, NeIn can be formatted so that its sample, negation prompt, and source image from COCO serve as the query image, query text, and retrieved image, respectively. \textit{For image-text matching}, VLMs can be trained to increase the cosine similarity between the COCO image and the negative instruction, while minimizing the similarity between the NeIn sample and the same caption, thus improving their ability to align images with textual negation.

\section{Photo Credits}

\begin{itemize}[noitemsep,topsep=0pt]
\itemsep0em 
    \item Goldendoodle Dog: \changeurlcolor{black}\href{https://www.instagram.com/aurixdoodle/}{instagram/aurixdoodle}
    \item Notre-Dame Cathedral Basilica of Saigon: \changeurlcolor{black}\href{https://www.indocinatours.it/viaggio-in-vietnam/ho-chi-minh-city-saigon/}{Indocinatours}
    \item Pizza: \changeurlcolor{black}\href{https://www.artanspizza.com/}{Artan's Pizza}
    \item Person and Dog: \changeurlcolor{black}\href{https://www.themanual.com/outdoors/going-outdoors-during-coronavirus-pandemic-self-quarantine-shelter-in-place/}{The Manual}
    \item Mocha Coffee and Book: \changeurlcolor{black}\href{https://www.mtzion.lib.il.us/mocha-mornings}{Mt Zion District Library}
    \item Beach Umbrella: \changeurlcolor{black}\href{https://www.nytimes.com/wirecutter/reviews/best-beach-umbrella-is-not-an-umbrella/}{Wirecutter}
    \item Candle: \changeurlcolor{black}\href{https://www.mcgeeandco.com/products/led-candle-lights-with-brass-clip-set-of-10}{McGee \& Co}
    \item Vegetable: \changeurlcolor{black}\href{https://www.themediterraneandish.com/how-to-make-vegetable-broth/}{The Mediterranean Dish}
    \item Cat and Broccoli: \changeurlcolor{black}\href{https://cats.com/can-cats-eat-broccoli}{Cats.com}
    \item Vietnamese Broken Rice Dish: \changeurlcolor{black}\href{https://sunhouse.com.vn/tu-van-mua-noi-com-dien/cach-nau-com-tam-bang-noi-com-dien.html}{Sunhouse}
\end{itemize}


\end{document}